\documentclass[conference]{IEEEtran}
\usepackage{cite}
\usepackage{amsmath,amssymb,amsfonts}
\usepackage{algorithmic}
\usepackage{graphicx}
\usepackage{textcomp}
\usepackage{xcolor}
\def\BibTeX{{\rm B\kern-.05em{\sc i\kern-.025em b}\kern-.08em
\\    T\kern-.1667em\lower.7ex\hbox{E}\kern-.125emX}}

\title{DeepTEGINN: Deep Learning Based Tools to Extract Graphs from Images of Neural Networks}

\author{%
\IEEEauthorblockN{Gustavo Borges Moreno e Mello}  
\IEEEauthorblockA{\textit{Dept. of Mech., Elec. and Chem. Engineering } \\
\textit{Oslo Metropolitan University}\\ 
Oslo, Norway\\
gustavo.mello@oslomet.no}\\

\IEEEauthorblockN{Sidney Pontes-Filho}
\IEEEauthorblockA{\textit{Dept. of Information Technology} \\
\textit{Oslo Metropolitan University}\\
Oslo, Norway \\
sidneyp@oslomet.no}\\

\IEEEauthorblockN{Ioanna Sandvig}
\IEEEauthorblockA{\textit{Dept. of Neuromedicine and Movement Science} \\
\textit{Norwegian University of Science and Technology}\\
Trondheim, Norway \\
ioanna.sandvig@ntnu.no}

\and
\IEEEauthorblockN{Vibeke Devold Valderhaug} 
\IEEEauthorblockA{\textit{Dept. of Neuromedicine and Movement Science}\\
\textit{Norwegian University of Science and Technology}\\
Trondheim, Norway\\
vibeke.d.valderhaug@ntnu.no}\\

\IEEEauthorblockN{Evi Zouganeli}
\IEEEauthorblockA{\textit{Dept. of Mech., Elec. and Chem. Engineering} \\
\textit{Oslo Metropolitan University}\\
Oslo, Norway \\
evi.zouganeli@oslomet.no}\\

\IEEEauthorblockN{Stefano Nichele}
\IEEEauthorblockA{\textit{Dept. of Information Technology} \\
\textit{Oslo Metropolitan University}\\
Oslo, Norway \\
stefano.nichele@oslomet.no}
}

\begin{document}
\maketitle
\begin{abstract}

In the brain, the structure of a network of neurons defines how these neurons implement the computations that underlie the mind and the behavior of animals and humans. Provided that we can describe the network of neurons as a graph, we can employ methods from graph theory to investigate its structure or use cellular automata to mathematically assess its function. Although, software for the analysis of graphs and cellular automata are widely available. Graph extraction from the image of networks of brain cells remains difficult. Nervous tissue is heterogeneous, and differences in anatomy may reflect relevant differences in function. Here we introduce a deep learning based toolbox to extracts graphs from images of brain tissue. This toolbox provides an easy-to-use framework allowing system neuroscientists to generate graphs based on images of brain tissue by combining methods from image processing, deep learning, and graph theory. The goals are to simplify the training and usage of deep learning methods for computer vision and facilitate its integration into graph extraction pipelines. In this way, the toolbox provides an alternative to the required laborious manual process of tracing, sorting and classifying. We expect to democratize the machine learning methods to a wider community of users beyond the computer vision experts and improve the time-efficiency of graph extraction from large brain image datasets, which may lead to further understanding of the human mind. \end{abstract}


\section{Introduction}
One of the goals of systems neuroscience is to obtain a mechanistic model that describes and explains how network of neurons in the brain implements perception, thought, and behavior. Because structure implements function in biology, an unavoidable step towards these mechanistic models is to describe the structural  connectivity of the network of neurons in the brain. Once a connectivity map (i.e., the description of the network) from a network of neurons is obtained, the it can be described as a graph or a cellular automata. This description may then be used as constraint to leverage methods from graph theory\cite{hassan1987review} to further analyze the network structure, or its functional complexity \cite{maccione2012multiscale}, which could ultimately contribute in three ways. First, by moving further the understanding on how to use biological substrates for computing  \cite{broersma2017computational,aaser2017towards}. Second, by improving computing methods in AI by supplying biologically derived network structures that can be used as reservoir networks \cite{nichele2017deep}. Finally by offering an mathematical abstraction of the biological system that can be used to compare cell cultures with genetic diseases with the healthy ones \cite{sandvig2018neuroplasticity}, providing ground for medical advancements. 

\begin{figure}
  \includegraphics[width=\linewidth]{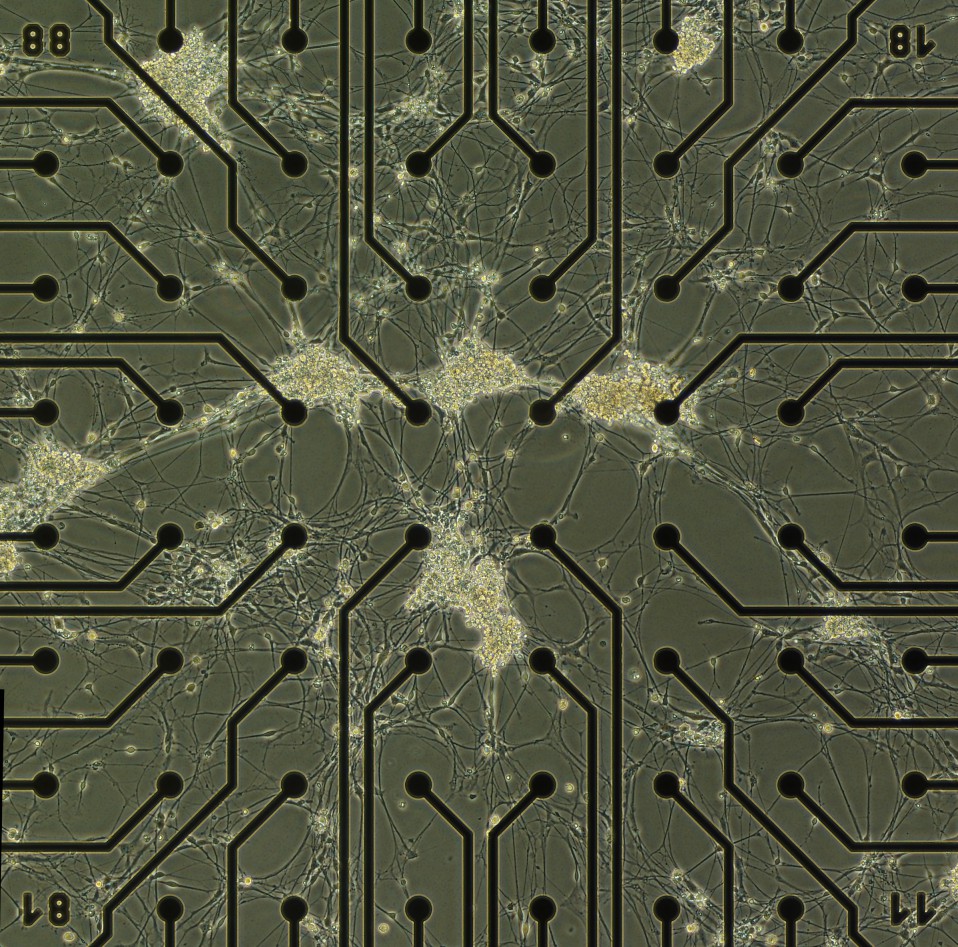}
  \caption{Raw image example as acquired from the microscope. The black lines ending in circle are "blind-spots" created by the multi-electrode array.}
  \label{fig:MEA}
\end{figure}

Nonetheless, obtaining these matrices of structural connections between the nodes of the nervous system is a challenging and cumbersome endeavour. Especially from microscopy images (see Figure \ref{fig:MEA}. Although some methods have been recently developed to automatize the process, they still rely on basic image processing steps that demand substantial time to find suitable parameters and curate the results \cite{dirnberger2015nefi}. Furthermore, these automatic methods are not very robust, and as an effect the gold standard approach still is to trace these connections manually. 

Additionally, and unlike other biological substrates from which networks can be extracted, the brain is constituted of billions of neurons, with diversified morphology, and a few orders of magnitude higher number of connections (i.e., synapses) \cite{kandel2000principles}. This additional complexity implies that different nodes in a graph network should have different properties and represent different cell types or structures in the brain. Because this structural information is critical to understand the brain, it is of utmost importance that automatic tools take them into account, which is not currently available to the best of our knowledge.

Furthermore, experimental constraints (e.g., multi-electrode arrays, patching pipettes) frequently obstruct the view of part of the image (black lines in Figure \ref{fig:MEA}), consequently preventing the connection of nodes that otherwise would be connected. Although these gaps can be easily handled through human intervention (i.e., by estimating the edges that connect two nodes), no simple image processing method can properly handle this problem as the data in the obstructed area is missing. To cope with this problem, one must be able to reconstruct the missing data by inferring how it would look like based on the surrounding area and what is typically known about the morphology.

Modern machine learning techniques that leverage the power of convolutional neural networks (ConvNets) may be used to automatize many of the steps mentioned before. In-painting algorithms can be used to estimate missing data caused by image obstruction, object detection algorithms can employed to locate and classify diverse structures in the brain, unsupervised segmentation algorithms can be leveraged to extract skeletonized versions of the image, just to name a few. This would allow for a comprehensive graph extraction from image in neuroscientific settings. The challenge regarding employing ConvNets is that due to its novelty and complexity, deep learning methods are not widely available to non-specialists in computer vision. Additionally, most of these methods require training, which by itself is generally poorly documented, preventing non-expert in computer vision from experimenting and benefiting from ConvNets in their neuroscientific research.

Motivated by the foregoing shortcomings, we present a deep learning based toolbox to extract graphs from images of brain tissue. This toolbox is a framework constituted by an extensible library of methods that can be integrated into a computer vision pipeline. The library is based on a combination of standard image processing algorithms available in OpenCV\cite{Braski2000Opencv} and SKimage \cite{van2014scikit}, and deep learning based methods for object detection, image/line segmentation, in-painting and style transfer implemented in Pytorch \cite{paszke2017automatic}. Additionally, the toolbox has a graphical user interface (GUI) that simplifies the steps of assembling the graph extraction pipeline, which includes the training of the supervised machine learning algorithms.

The main contribution of this paper is to democratize deep learning based methods for computer vision to the neuroscientific community through a reusable, flexible and scalable tool. Through this toolbox, we hope to make deep learning methods more widely accessible to neuroscientists.

\section{Image acquisition}
The images were prepared as follows: Human cortical neural networks were differentiated and matured from iPSC-derived NSCs (ax0019, Axol bioscience), and fluorescently labelled using a two-color LIVE/DEAD viability/cytotoxicity kit (MP03224, Invitrogen). 0.8ul Ethidium homodimer-1 (2mM in DMSO/H2O 1:4) and 0,4ul Calcein AM (4mM in anhydrous DMSO) was diluted in 2ml PBS and applied to the neural networks for 15 minutes in 37\textdegree C. The former produces an intense red fluorescence in dead or dying cells, while the latter produces an green fluorescence in live cells. The fluorescently labelled neural networks were imaged with a 10X objective using a automated EVOS 2 fluorescence microscope.

Each multi-electrode array (MEA) cell culture chamber was briefly sterilized using ethanol, washed with water, UV-treated over-night, and hydrophilized by application of foetal bovine serum for 30-60 minutes at room temperature. The surface was subsequently double-coated using poly-L-ornithine (0,01\%) and laminin. The appropriate neuronal cell culture media were heated to 37\textdegree C and used to create a single-cell suspension, from which 100.000 cells were seeded directly onto the electrode area of each MEA in a dropwise manner. For some cultures, a feeder-layer of astrocytes (5000 per MEA) was first established, upon which 50 000 neuronal cells were seeded onto. The MEA neuronal cultures were kept in a standard humidified air incubator (5\% CO2, 20\%O2, 37\textdegree C), and 50\% of the media were changed every 2-3 days. Phase contrast images were acquired at various stages of neuronal differentiation and maturation on the MEAs using the laboratory light microscope Carl Zeiss Axiovert 25 with 5 and 10X objectives.

\section{The pipeline}
The kernel of the toolbox is the graph extraction pipeline. It enables visualization, correction and analysis of the structures depicted in the input image. This pipeline, is constituted by an ordered sequence of methods, which will output a graph representation of the network from the input image. Additionally, some of the steps of the pipeline sequence require preparation (i.e. training). The obtained graph provides weights, edge lengths and node type, which should reflect anatomical structure.

The default pipeline combines the following steps: pre-processing, structure detection, segmentation, thinning, graph extraction, graph pruning, and training. What follows is a high-level description of the steps.

\subsection{Pre-processing}
\begin{figure*}[ht]
  \includegraphics[width=\textwidth]{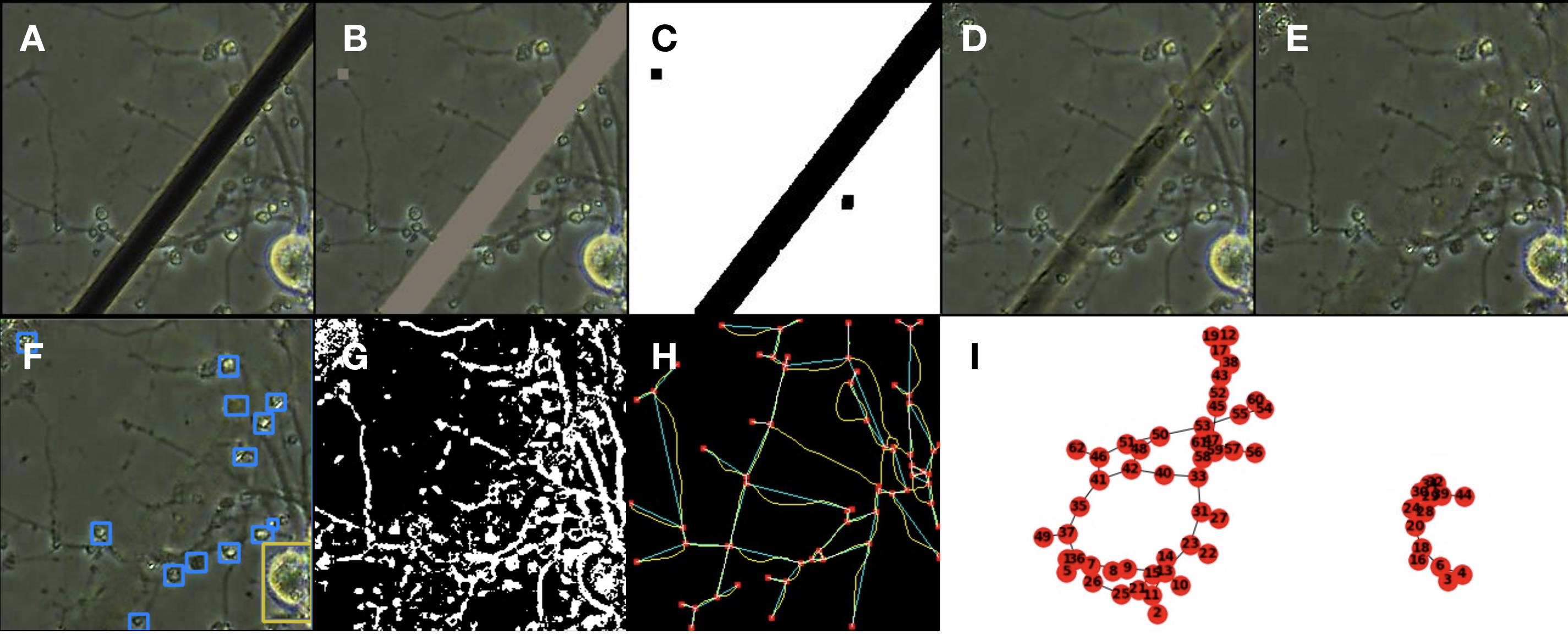}
  \caption{Graph extraction example. A- 256 x 256 Crop of a raw image as acquired from the microscope. B- Segmentation of the electrode area in grey. C- Mask of the area to be in-painted. D- Intermediate step in the inpainting process. E- End result of the in-painting algorithm. F- Cells identified by type with yolov3, blue boxes are neurons, yellow box is a cluster of cells, astrocytes are omitted. G- Intermediate step of the structure segmentation. H- Graph extraction example, the red represents the nodes, the blue are the edges connecting the nodes and the yellow is the thinned (skeletonized) version of the network identified. I- Graph output describing the network in H.}
  \label{fig:pipeline}
\end{figure*}

Pre-processing involves doing image transformations that allows the subsequent algorithms to perform more robustly. There is a set of image processing steps that may be employed interchangeably. Most of them are standard image transformations like color space change and filtering (including sharpening and blurring), widely available through OpenCV and SKimage libraries. Additionally, deep learning based algorithms were included, namely: style transfer \cite{gatys2015neural} and in-painting\cite{liu2018image}. These two methods rely on VGG16 networks pre-trained on ImageNet dataset\cite{simonyan2014very} and must be further fine-tuned to properly work with the dataset from the experiments (see training). The addition of in-painting(Figure \ref{fig:pipeline} D) and style transfer (Figure \ref{fig:pipeline} E)  enables coping with data lost caused by obstruction of the field of view during experiments, and differences in imaging settings respectively (See black marks in Figure \ref{fig:MEA}).

\subsection{Structure detection}
In order to detect nodes that belong to different cell types, the object detection algorithm, yolov3 (You Only Look Once, version 3.0) \cite{redmon2018yolov3}, was included in the pipeline. This method operates by detecting combinations of spatial features in the image, locating their position and area, and classifying them under a predefined category (namely: astrocytes, neurons and clusters of neurons; see Figure \ref{fig:pipeline} F) with an explicit probability. This algorithm requires training, and substantial amount of data to be trained (see training). The center of these detected areas is used by later steps in the pipeline to discriminate synaptic nodes from cell body nodes.

\subsection{Segmentation}
We provide two interchangeable avenues for segmentation. Guided watershed \cite{osma2007improved} and W-Net \cite{xia2017w}. We noticed that depending on the characteristics of input image these algorithms perform the best. The main goal of this step is to separate the structures that compose the network from everything else and compose a mask. 

\subsection{Thinning}
The next step is to skeletonize the mask so no pixel in the mask has two or more neighbor pixels that belongs to the mask and are neighbor to each other (see Figure \ref{fig:pipeline} H, yellow lines). To do that, we implemented the improved Zhang-Suen Thinning algorithm \cite{chen2012improved}. This method was chosen because it produced less artifacts in the intersection of lines (blobs and missing pixels).

\subsection{Graph extraction}
Once obtained the skeletonized image we then detect the positions of nodes and the edges that connect them so we can create a graph. The graph is generated through the NetworkX \cite{hagberg2008exploring} library. To detect the nodes, we filter the image with series of 3x3 filters. Each filter represent one possible scenario for a node where the center pixel belongs to the filter and 1 or at least 3 other neighbours also belong to it and are not neighbours to each other. This guarantees that intersection nodes and end-of-the-line nodes are contemplated, but that points that belongs to lines are ignored.

Edges between nodes are detected by the following steps. Firstly, the skeleton is segmented in edges by removing the node pixels from the skeletonized mask, each one of these edges has its own label automatically defined as 1 to the number of available edges. Secondly, these segmented edges are dilated. Thirdly, the edges that overlap with two nodes are added to the graph as a bidirectional edge. Finally, the overlapping segmented edge is removed from the set of possible edges. The process repeats until no edges are left (see Figure \ref{fig:pipeline} H blue lines).

If the Structure Detection step has been executed successfully, the nodes which are closest to the center of the  regions of interest generated by the object detection algorithm will acquire the category of the identified object (eg., neuron, cluster).

This type of representation of the network in graph is analogous to a connectivity estimation, which is highly relevant in neuroscience to infer functionality. Hence the graph extraction method can contribute to connectivity analysis as performed by Maccione et al. \cite{maccione2012multiscale} and Ullo et al. \cite{10.3389/fnana.2014.00137}, but without the cumbersome and time-demanding step of extracting the connectivity map manually.

\subsection{Graph pruning}
As the last step, it is given to the user the opportunity to edit the graph extracted by removing or adding new edges, tracing new edges between them and assigning properties to each node. 

\subsection{Training}
In order to use the methods that depend on supervised learning, the user has to provide first a set of good examples so the algorithm can be properly trained. This process is usually poorly documented and the data format that the algorithm should receive is usually obscure. We make this step explicit, by declaring exactly what should be the data format, and providing a simple interface that the users can use to generate the training data themselves and train the model.

\subsection{Training in-painting}
To train the in-painting\cite{liu2018image} network, we need a set of ground truth images, and a set of masks that would match in shape and size the typical artifact obstructing the original image. To generate this data, we segmented the dark areas of original raw image using a simple threshold. We dilated the segmented areas with a 5x5 circular kernel to include the edges of the artifact areas. We randomly cropped the mask image in images of 256x256 pixels. The images in which 1/4th of the area was occupied by the electrode mask were selected for the mask pool. Each mask was then copied 35 times and rotated cumulatively by 10 degrees until we had 36 versions of the same mask in all orientations.
To extract the ground-truth images, we cropped patches of 256x256 pixels from the original image where no pixel in the patch overlapped with the cordinates of a pixel belonging to a mask. To expand the dataset, the selected patches were flipped and rotated 90, 180 and 270 degrees.

\subsection{Training object detection}
To train the Yolov3, we defined regions of interests (ROIs) by drawing a bounding boxes from a subset of patches (100 images, randomly picked). This ROIs are constrained by the vertical and horizontal coordinates of its centroid and its height and with as ratios of the original image. Each ROI is labeled as an instance of a class of objects. 

In our dataset, we defined three labeled structures: neurons, astrocytes and cluster of neurons. Only neurons and cluster of neurons were relevant for the graph extraction, thus only these two labels are displayed. The labeling of astrocytes was required to prevent falsely detecting astrocytes as neurons. Once training data is available, training follows by pointing the location of the data in the storage unit and running the training function.

Because the amount of data was limited. The network was pre-trained with the COCO dataset\cite{lin2014microsoft} to learn  and then fine-tuned and cross-validated using the labeled images. Our implementation of Yolov3 operates by predicting 3 boxes in 2 different scales. Thus, the tensor is N x N x[2*(4+1+3)], where is for the 4 bounding box offsets, 1 for objectness prediction, and 3 is for the class predictions. Furthermore we chose 6 clusters in the k-means algorithm to establish our binding box priors. In our dataset the 6 clusters were (7x9), (15x16), (22x19), (31x32), (55 x 49), (89x91).

The training progress was displayed on every set of epochs, which could defined by the user, and it could be interrupted at any time. The set of weights with the smaller error was highlighted to facilitate the use of the pipeline.

\section{Discussion}
Many solutions exist to extract network graphs from images, including some generic flexible tools. But these tools assume that the network to be extracted is homogeneous (i.e., all the nodes are equal). This is a major problem in neuroscience because biological neurons form highly heterogeneous networks. In particular, the tools available cannot account for the difference between neurons and synapses as nodes. Additionally, they cannot account for differences between cell types (i.e., neurons vs. glia). This is a major source of error for describing a network. The abundance of false positives can lead to a description that is much bigger and dense, hence increasing the level of complexity which by itself increases the challenge of analysis. Our toolbox circumvent this problem by integrating machine learning methods into an easy to use pipeline to extract graphs from network of neurons. Because the algorithm detects objects by category, further developments may be implemented to extract sub-populations of neurons and include more cell types. One possible avenue is to develop a specific dataset for brain cell-type detection, which is currently unavailable in the best of our knowledge. 

A second major challenge in the path of automatizing graph extraction from images of cultured cells is that these images often come with major artifacts (e.g., electrodes, pipettes, objects that obstruct the view). We eliminate these artifacts by combining in-painting techniques and style transfer through deep learning methodologies. Although not perfect, we demonstrate that both techniques can provide qualitatively satisfactory results. Allowing to reconstruct a plausible network, despite the artifact. One further point of development could be to apply techniques to improve the resolution, as it may increase the performance of in-painting and style transfer techniques.

We anticipate that this toolbox will  enable neuroscientists to extract graphs from network of neurons in a more time-efficient way and consequently contribute in the pursue of the understanding of perception, intelligence and behavior.

\section*{Acknowledgment}
This work was supported by Norwegian Research Council SOCRATES project (grant number 270961) and received internal support as a lighthouse project in Computer Vision from the Faculty of Technology, Art and Design (TKD) at Oslo Metropolitan University, Norway.

\section*{Repository}
Code and example data can be found in the following repository:
https://github.com/gmorenomello/deepteginn

\bibliographystyle{ieeetr}
\bibliography{bibliography.bib}

\end{document}